# EMOCPD: Efficient Attention-based Models for Computational Protein Design Using Amino Acid Microenvironment

Xiaoqi Ling[1,+], Cheng Cai[1,+], Demin Kong[2], Zhisheng Wei[2], Jing Wu[2], Lei Wang[2,*], Zhaohong Deng[1,*]
1. The School of Artificial Intelligence and Computer Science, Jiangnan University, Wuxi 214122, China
2. The School of Biotechnology, Jiangnan University, Wuxi 214122, China.
+ Equal contributions;
* Corresponding author: dengzhaohong@jiangnan.edu.cn (Zhaohong Deng), leiwang_enzyme@jiangnan.edu.cn (Lei Wang)

**Abstract**

Computational protein design (CPD) refers to the use of computational methods to design proteins. Traditional methods relying on energy functions and heuristic algorithms for sequence design are inefficient and do not meet the demands of the big data era in biomolecules, with their accuracy limited by the energy functions and search algorithms. Existing deep learning methods are constrained by the learning capabilities of the networks, failing to extract effective information from sparse protein structures, which limits the accuracy of protein design. To address these shortcomings, we developed an Efficient attention-based Models for Computational Protein Design using amino acid microenvironment (EMOCPD). It aims to predict the category of each amino acid in a protein by analyzing the three-dimensional atomic environment surrounding the amino acids, and optimize the protein based on the predicted high-probability potential amino acid categories. EMOCPD employs a multi-head attention mechanism to focus on important features in the sparse protein microenvironment and utilizes an inverse residual structure to optimize the network architecture. The proposed EMOCPD achieves over 80% accuracy on the training set and 68.33% and 62.32% accuracy on two independent test sets, respectively, surpassing the best comparative methods by over 10%. In protein design, the thermal stability and protein expression of the predicted mutants from EMOCPD show significant improvements compared to the wild type, effectively validating EMOCPD's potential in designing superior proteins. Furthermore, the predictions of EMOCPD are influenced positively, negatively, or have minimal impact based on the content of the 20 amino acids, categorizing amino acids as positive, negative, or neutral. Research findings indicate that EMOCPD is more suitable for designing proteins with lower contents of negative amino acids.

## 1 Introduction

Computational Protein Design (CPD) refers to a class of methods that utilize computer algorithms and models to predict and optimize protein structures and functions, aiming to discover better protein structures for specific biological or application purposes. Over the past few decades, CPD has achieved significant success in various fields, including enzyme engineering[1-3], vaccine design[4, 5], antibody design[6, 7], membrane protein design[8, 9], and protein interactions[10-12]. For example, recent work published in Nature by Hongyuan Lu et al[13]. successfully designed a PET enzyme with high stability and hydrolytic activity, capable of hydrolyzing 51 types of PET plastics in just one week, potentially addressing the global plastic waste issue. Ongoing research indicates that CPD can benefit society in health, medicine, environment, and chemical engineering. Despite the notable advancements, effectively designing proteins with specific functions remains a considerable challenge.

Traditional CPD research primarily relies on Anfinsen's folding thermodynamics hypothesis[15]. The core idea of this hypothesis is that, under natural conditions, proteins always fold into the conformation with the lowest free energy. This means that a protein's three-dimensional structure is entirely determined by its amino acid sequence[16-17]; each protein sequence corresponds to a unique three-dimensional structure, although similar structures may correspond to multiple sequences. Traditional CPD methods assume that the structure with the lowest free energy is the optimal one. Therefore, the focus of traditional CPD is on systematically substituting amino acids in the protein to find the sequence that minimizes free energy. The primary concerns of traditional CPD are the design of free energy calculation functions and the search for the lowest energy amino acid sequences. Widely used energy function-based CPD methods include RosettaDesign[18], EvoDesign[19], and SCUBA[20]. These methods have developed their energy functions and utilize heuristic algorithms to search the vast amino acid sequence space for the lowest energy sequences. However, energy function-based methods are often limited by the design of their energy functions and can easily become trapped in local optima, leading to proteins that do not achieve the target structure and resulting in a low success rate for protein design experiments[21].

In recent years, deep learning has made significant breakthroughs across various fields, including computer vision[22-24], natural language processing[25-27], computational chemistry[28, 29], and computational biology[30-32]. New methods using deep learning for protein design have emerged in the CPD domain. These methods aim to learn the correspondence between amino acid residues and their surrounding structural features based on the local chemical environment of the residues. By leveraging these relationships, they modify certain amino acids in the protein without disrupting the protein backbone, thereby achieving protein design. Depending on the network architecture used, these methods can be categorized into those based on Multi-Layer Perceptron (MLP), Convolutional Neural Networks (CNN), and Graph Convolutional Networks (GCN).

Representative MLP-based methods include SPIN[33] and SPIN2[34]. SPIN inputs local geometric features and global energy features of amino acids into a two-layer neural network to predict amino acid types, achieving an accuracy of 30.7% on a test dataset consisting of 500 proteins (TS500). SPIN2 builds on SPIN by adding more local geometric features and improving the neural network structure, increasing the prediction accuracy to 34.6%. MLP-based methods are limited by the calculation of handcrafted features; the more comprehensive the handcrafted feature design, the better the classification performance. However, handcrafted features often have limitations, causing MLP-based prediction accuracy to reach a bottleneck.

Representative CNN-based methods include SPROF[35], ProDCoNN[36], and DenseCPD[37]. The SPROF method uses a residue adjacency matrix as input, establishing a hybrid neural network model that combines CNN and biLSTM, which improves amino acid prediction accuracy to 39.8%. ProDCoNN employs a 3D CNN based on three-dimensional environmental features composed of $N$、$C_\alpha$、$C$、$O$ atoms for amino acid classification, achieving a prediction accuracy of 42.2% on a test set of 500 proteins. DenseCPD uses the DenseNet[38] architecture for its prediction model, achieving an accuracy of 54.5% on the TS500 test set based on three-dimensional environmental features composed of $N$、$C_\alpha$、$C$、$O$ and $C_\beta$ atoms. CNN-based methods primarily model the three-dimensional environmental features of amino acids. However, the sparse distribution of atoms in three-dimensional space can hinder the convergence of neural network training. To overcome atom sparsity and accelerate network convergence, CNN-based methods typically apply Gaussian functions to blur the three-dimensional environmental features, which results in some loss of information and affects the final prediction performance.

Representative GCN-based methods include GraphTrans[39] and ProteinMPNN[40]. GraphTrans constructs a graph for the target protein's amino acids, encoding the nodes as types of amino acid residues and the edges based on the distances, directions, and origin vectors of the residues, followed by training a Transformer[25] encoder-decoder model. GraphTrans achieves a prediction accuracy of 49.6% on its test set and 44.7% on the TS500 dataset. ProteinMPNN employs two three-layer message-passing networks as the encoder and decoder. This method uses the encoder to extract node and edge features from the distances of amino acid residues $N$、$C_\alpha$、$C$、$O$ and $C_\beta$ atoms., then utilizes the decoder to convert these features into amino acid probabilities, ultimately achieving an accuracy of 58.1% on the TS500 dataset. GCN-based methods establish graphs of the atomic environment, which helps mitigate the sparsity of atoms in space but neglects the three-dimensional positional information of the atoms.

To address the limitations of existing deep learning approaches in protein design and to further enhance the accuracy of CPD, this paper proposes a new neural network model, EMOCPD, for amino acid recognition and protein design. EMOCPD incorporates a multi-head attention mechanism to learn important features from sparse three-dimensional microenvironments of amino acids, and it introduces the Inverted Residual Mobile Block (iRMB) from the EMO model[41] to enhance the model's learning performance. Our experimental results demonstrate a significant improvement in prediction accuracy compared to existing deep learning-based CPD methods. The amino acid prediction results obtained using EMOCPD facilitate the design of proteins with superior properties.

## 2 Materials and Methods

### 2.1 Dataset

This study utilizes a self-constructed training dataset, as well as two commonly used testing datasets: TS50 [36] and TS500 [37].

#### 2.1.1 Training Dataset

To construct the training dataset, we selected proteins from the PDB database. The screening process was as follows: First, to avoid over-sampling similar protein structures, which could lead to model overfitting, we filtered proteins with a sequence similarity not exceeding 50%. Second, to ensure high-quality proteins, we excluded those with a resolution less than 2.5 Å. Finally, we removed proteins that were included in the TS50 and TS500 testing datasets. After the screening, a total of 20,118 proteins were selected for training, with 298 proteins used for validating the model's performance during training. Additionally, to avoid excessive sampling of amino acids from large proteins (number of amino acids > 200), we referenced literature [14] and implemented the following approach: Let nnn be the number of amino acids in a protein. If nnn exceeds 200, we randomly selected 100 amino acids; otherwise, we randomly selected all nnn amino acids. Following this processing, our constructed training set for amino acid recognition contains 1,605,000 amino acids, of which 8,000 are used for validating model performance during training. The filtered protein files used in this study can be accessed from the PDB database [42], and the PDB IDs of each protein are listed in Appendix 1.

#### 2.1.2 Testing Dataset

The TS50 and TS500 datasets have been used in numerous studies [33, 34, 35, 36, 37] to evaluate the performance of CPD methods. Therefore, we also employed these two datasets for performance testing and compared them with other representative CPD methods. The TS50 dataset comprises 50 single-chain protein structures, totaling 6,861 amino acids. The TS500 dataset consists of 500 proteins (including 887 protein chains), encompassing a total of 216,429 amino acids. The protein files for both datasets can be obtained from the PDB database, and the PDB IDs and chain numbers of the three-dimensional structures are provided in Appendices 2 and 3.

## 2.2 EMOCPD Model

The proposed protein design model, as illustrated in Figure 1, consists of several key modules: a) Atom Feature Construction Module; b) Microenvironment Grid Construction Module; c) EMO Deep Learning Module; d) MLP Classifier Module. The function of the Atom Feature Construction Module is to encode protein atoms into atomic feature vectors. The Microenvironment Grid Module is responsible for constructing a grid from the atomic feature vectors of the protein to represent the amino acid microenvironment. The EMO Deep Learning Module utilizes the EMO model to learn deep features from the amino acid microenvironment. Finally, the MLP Classifier Module uses these deep features to predict the amino acid category that best fits the amino acid microenvironment. Each module will be described in further detail in the subsequent sections.

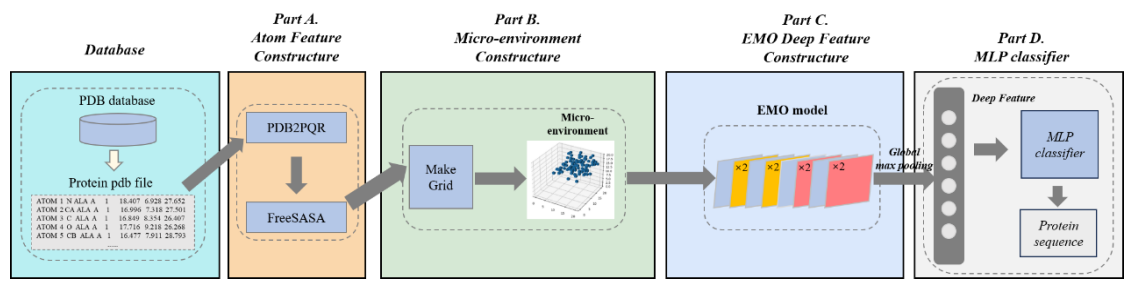

**Figure 1** Schematic diagram of the overall framework of EMOCPD. Part A: Atom Feature Construction Module; Part B: Microenvironment Grid Construction Module; Part C: EMO Deep Learning Module; Part D: MLP Classifier Module.

### 2.2.1 Atom Feature Construction Module

To compute the features of protein atoms for establishing the environmental representation of amino acids, we performed the following steps: 1) Hydrogen atoms were added to the protein in the PDB file; 2) The free charge (FC) of each atom was calculated for the protein with added hydrogen atoms; 3) The solvent-accessible surface area (SASA) of each atom was calculated. We used the bioinformatics program PDB2PQR[43] to add hydrogen atoms and compute the FC, while FreeSASA[44] was used to calculate the SASA of each atom. After these preparations, each atom was encoded as a 7-dimensional vector, where the first five dimensions represent the one-hot encoding of the atom type, and the last two dimensions represent the FC and SASA of the atom.

### 2.2.2 Microenvironment Grid Construction Module

This study transforms the amino acid recognition problem into an image classification problem. To achieve this, the three-dimensional structure of the amino acids is converted into a 3D voxel grid. A 3D grid is established with the amino acid $C_\beta$ as the center, where the direction of the amino acid $N - C_\alpha$ bonds serves as the x-axis, the direction perpendicular to the $N - C_\alpha - C$ plane and positively correlated with the $C_\alpha - C_\beta$ bond direction serves as the z-axis, and the direction orthogonal to the xz-plane serves as the y-axis. The grid size is 20Å × 20Å × 20Å, with each cubic unit measuring 1Å × 1Å × 1Å. The atom $C_\beta$ is selected as the center to maximize the capture of the environment surrounding the amino acid side chain. For glycine, which has no $C_\beta$ atoms, its coordinates are calculated from the $N - C_\alpha$ lengths of and $N - C_\alpha - C_\beta$ bond angles of the other amino acids. After removing the side chain (R-group) atoms of the central amino acid, each grid unit is encoded as the sum of the feature vectors of the atoms it contains. Ultimately, each amino acid is represented as a $7 \times 20 \times 20 \times 20$ voxel grid, with the label corresponding to the amino acid type.

### 2.2.3 EMO Deep Learning Module

The overall framework of the proposed EMO model is shown in Figure 2, which consists of one Stem, four iRMB modules, four MHSA-iRMB modules, and three DownSample modules. The specific parameters for each module are provided in the figure. Figure 3 illustrates the various modules of the EMO model, with detailed descriptions to follow in subsequent sections.

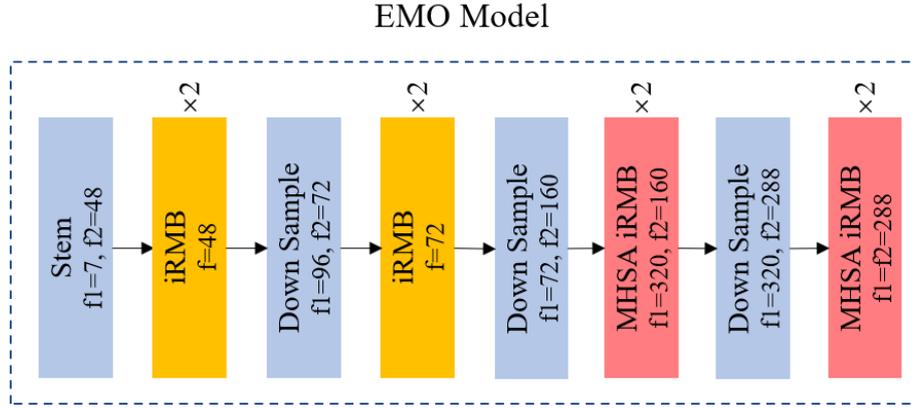

**Figure 2** Schematic diagram of the EMO model.

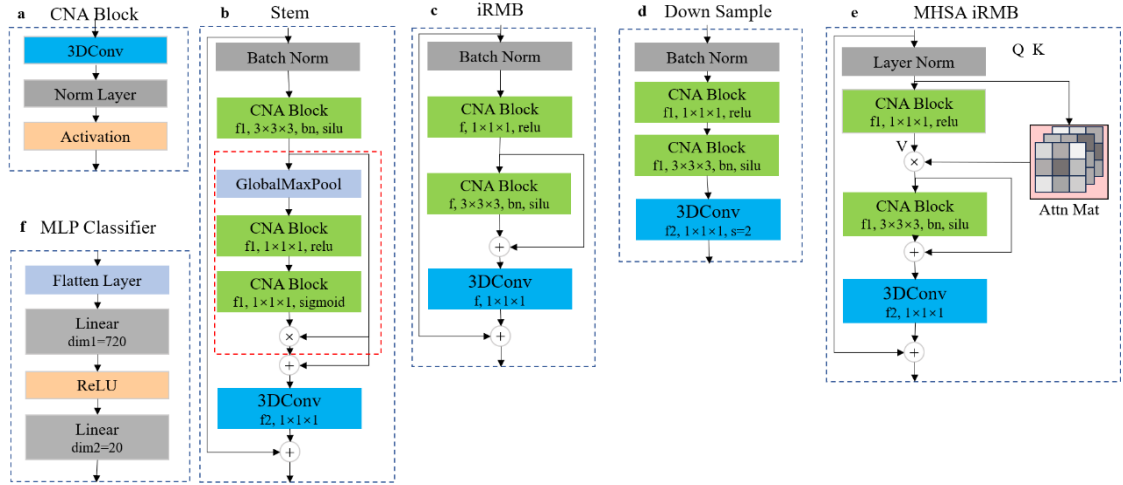

**Figure 3** Schematic diagram of the structural composition of each module in the EMOCPD. (a) CNA component, (b) Stem module, (c) iRMB module, (d) Down Sample module, (e) MHSA iRMB module, (f) MLP classifier module. The characters appearing below the CNA module in (b) to (e) represent the number of convolution kernels, kernel size, normalization layer used, and activation function layer parameters for the 3D convolution within the CNA module, respectively.

### 1) CNA Component

The CNA (ConvLayer, NormLayer, Active) component is an extension of the CBA (ConvLayer, BatchNormLayer, Active) module from ResNet[45]. It allows for the selection of the desired normalization layer and activation function based on specific requirements. As this component is an important part of modules such as Stem in the proposed model (Figure 3(b)–(e)), it will be introduced first.

As shown in Figure 3(a), the CNA component consists of a 3D convolution layer, a normalization layer, and an activation function layer. Given an input $A_{in}$, a CNA component can be represented as:

$$f_{CNA}(A_{in}) = f_{active}\left(f_{norm}(f_{conv}(A_{in}))\right) \qquad (1)$$

Here, $f_{conv}$ denotes the 3D convolution layer, while $f_{norm}$ refers to the normalization process, and $f_{active}$ signifies the activation function. Common normalization layers include batch normalization (bn) and layer normalization (ln), while common activation functions include $relu, sigmoid$, and $silu$. The CNA component will be consistently referred to as $f_{CNA}(\cdot)$ in subsequent discussions.

**2) Stem Module**

Figure 3(b) illustrates the Stem module of EMOCPD, which includes three CNA modules, a batch normalization layer, a global max pooling layer, a 3D convolution projection layer, and three residual connections. The part outlined by the red dashed box can be considered as an SE module (Squeeze-and-Excitation) with a scaling factor of 1[46]. Given an input $A_{in}$, the SE module can be expressed as:

$$f_{SE}(A_{in}) = f_{CNA}(f_{CNA}(f_{GMP}(A_{in}))) \qquad (2)$$

The equation $f_{GMP}$ represents the global max pooling layer. Let $B = f_{CNA}(f_{bn}(A_{in}))$. Therefore, the Stem module can be expressed as:

$$f_{Stem}(A_{in}) = A_{in} + f_{conv}(B + f_{SE}(B) \cdot B) \qquad (3)$$

In the equation, $f_{SE}$ represents the SE module, and $f_{bn}$ denotes the Batch Normalization layer. The first CNA module in the Stem module utilizes a Batch Normalization layer for normalization, with the activation function set to *silu*. The number of convolutional filters is f1, and the filter size is 3 × 3 × 3. In the SE module, the two CNA modules use *relu* and *Sigmoid* activation functions, respectively, with both having f1 filters and a filter size of 1 × 1 × 1. $f_{conv}$ represents the 3D convolutional layer, with a filter size of 1 × 1 × 1 and f2 filters. The specific values for the parameters f1 and f2 in the Stem module of the EMO model are detailed in Figure 2.

**3) iRMB Module**

Figure 3(c) illustrates the iRMB module, which consists of a batch normalization layer, two CNA modules, a 3D convolutional layer, and two residual connections. Given an input $A_{in}$, let $B = f_{CNA}(f_{bn}(A_{in}))$. Then, an iRMB module can be represented as:

$$f_{iRMB}(A_{in}) = A_{in} + f_{conv}(f_{CNA}(B) + B) \qquad (4)$$

The two CNA modules in the iRMB module utilize *relu* and *silu* activation functions, with convolution kernel sizes of 1 × 1 × 1 and 3 × 3 × 3, respectively. The first CNA module does not include a normalization layer, while the second one employs batch normalization. The kernel size of the 3D convolutional layer is 1 × 1 × 1, and the number of kernels for both CNA modules and the 3D convolutional layer is denoted as f. The specific values for the parameter f in the iRMB module of the EMO model are detailed in Figure 2.

**4) Down Sample Module**

Figure 3(d) depicts the down sample module, which consists of a batch normalization layer, two CNA modules, and a 3D convolutional layer. Given input $A_{in}$, a down sample module can be represented as follows:

$$f_{down}(A_{in}) = f_{conv}\left(f_{CNA}\left(f_{CNA}(f_{bn}(A_{in}))\right)\right) \qquad (5)$$

The two CNA modules in the down sample module utilize *relu* and *silu* activation functions, both with f1 filters. The kernel sizes are 1 × 1 × 1 and 3 × 3 × 3, respectively. The first CNA module does not include a normalization layer, while the second employs batch normalization. The 3D convolutional projection layer has a kernel size of 1 × 1 × 1, with f2 filters and a stride of 2. The down sample module effectively reduces the feature scale by half. The specific values for the parameters f1 and f2 in the Down Sample module of the EMO model are detailed in Figure 2.

5) **MHSA iRMB Module**

Figure 3(e) illustrates the MHSA iRMB module, which integrates a multi-head self-attention mechanism into the iRMB module. This module comprises a batch normalization layer, two CNA modules, a multi-head self-attention mechanism module, a 3D convolution layer, and two residual connections. The multi-head self-attention mechanism operates on the features generated by the first CNA module. Given the input $A_{in}$, the multi-head attention mechanism can be represented as follows:

$$f_{MHSA}(A_{in}) = Concat(H_1, \ldots, H_h)W_o \tag{6}$$

In the equation, $W_o$ represents the weighted matrix for the multi-head attention scores, and $H_i$ denotes the self-attention score for the i-th head, which can be expressed as:

$$H_i = Attention(Q_i, K_i) = SoftMax\left(\frac{Q_i K_i^T}{\sqrt{d_{K_i}}}\right) \tag{7}$$

In this context, $Q_i$ and $K_i$ refer to the query and key vectors for the i-th self-attention, while $d_{K_i}$ denotes the dimension of the i-th key vector, and $V$ represents the value matrix. In the MHSA iRMB module, the calculations for $Q_i$, $K_i$, and $V$ are given by: $Q_i = f_{CNA}^{q_i}(f_{ln}(A_{in}))$, $K_i = f_{CNA}^{k_i}(f_{ln}(A_{in}))$, and $V = f_{CNA}^{v}(f_{ln}(A_{in}))$. Let $Att = f_{MHSA}(A_{in}) \cdot V$ represent the attention mechanism, then the MHSA iRMB module can be expressed as:

$$f_{MHSA\_iRMB}(A_{in}) = f_{ln}(A_{in}) + f_{conv}(f_{CNA}(Att) + Att) \tag{8}$$

In the equation, $f_{ln}$ represents the layer normalization layer. In the MHSA iRMB module, the two CNA components utilize the *relu* and *silu* activation functions, respectively, with both having the same number of filters f1. The filter sizes are 1 × 1 × 1 and 3 × 3 × 3. The first CNA module does not include a normalization layer, while the second uses batch normalization. The 3D convolution projection layer has a filter size of 1 × 1 × 1 and a number of filters f2. The specific values for the parameters f1 and f2 in the MHSA iRMB module of the EMO model can be found in Figure 2.

2.2.4 **MLP Classifier Module**

Figure 3(f) illustrates the MLP classifier module, which consists of a global max pooling layer, a flatten layer, two linear layers, and *relu* activation function layers in between. Given the input $A_{in}$, the MLP classifier can be represented as follows:

$$f_{MLP\_classifier}(A_{in}) = f_{linear}\left(f_{relu}\left(f_{linear}\left(f_{flatten}(f_{GMP}(A_{in}))\right)\right)\right) \tag{9}$$

In the equation, $f_{linear}$ denotes the linear layer, $f_{relu}$ represents the relu activation function layer, and $f_{flatten}$ indicates the flatten layer. The hidden node counts for the two linear layers are 720 and 20, respectively.

**2.2.5 Model Training**

For the proposed EMOCPD model, we implemented and optimized it using the PyTorch[47] framework. The input is a 7 × 20 × 20 × 20 voxel grid representing amino acids, and the output is a 20-dimensional vector indicating the predicted amino acid labels. During training, the loss function used is the cross-entropy loss, and the optimizer is the Adam optimizer[48], with a learning rate set to $10^{-5}$ and a weight decay coefficient of $10^{-3}$; the remaining parameters are set to PyTorch's default values. The training is configured for 8 epochs, with 10,700 steps per epoch, each containing 150 samples.

# 3 Experimental Results

To evaluate the performance of the proposed method, experimental studies were conducted from three aspects: 1) experimental analysis on the training set; 2) comparison with existing representative methods on the test set; 3) biological experimental validation of protein design using EMOCPD.

## 3.1 Evaluation Metrics

In this paper, we used four metrics—accuracy (Acc), recall (Rec), precision (Pre), and F1 score—to evaluate the performance of our proposed method. Accuracy is used to assess the overall prediction performance for the 20 amino acids, while recall, precision, and F1 score are used to evaluate the prediction performance for each individual amino acid. The definitions of the four metrics are as follows:

$$ACC = \frac{TP+TN}{TP+TN+FP+FN} \tag{10}$$

$$Rec = \frac{TP}{TP+FN} \tag{11}$$

$$Pre = \frac{TP}{TP+FP} \tag{12}$$

$$F1 = \frac{2TP}{2TP+FP+FN} \tag{13}$$

In the equations, TP, TN, FP, and FN represent the counts of true positives, true negatives, false positives, and false negatives, respectively. Accuracy refers to the ratio of correctly classified samples to the total number of samples; recall indicates the proportion of actual positive samples that are correctly identified as positive; precision refers to the proportion of correctly predicted positive results among those predicted as positive; and the F1 score is the harmonic mean of recall and precision, allowing for a balanced assessment of classification performance.

Additionally, we also utilized TOP-K accuracy. TOP-K accuracy is a commonly used metric in image classification, defined as the proportion of correct labels included in the top K predicted results ranked by their probabilities in descending order.

## 3.2 Training Dataset Results

Figure 4 illustrates the change in training and validation accuracy over the number of iterations, while Figure 5 presents the confusion matrix for the predicted results of the 20 amino acids. From Figure 4, it can be seen that after 8 epochs and a total of 85,600 iterations, our model achieves an accuracy exceeding 80% on the training samples, with an upward trend still evident as the number of iterations increases. However, the accuracy on the validation samples stabilizes after approximately 60,000 iterations, reaching a peak of around 63%. This indicates a significant risk of overfitting if training continues, leading us to decide to stop training after the 8th epoch. As shown in Figure 5, our model achieves an accuracy of over 40% for all amino acids except glutamine (GLN, Q), with particularly high accuracies exceeding 90% for proline (PRO, P), glycine (GLY, G), and cysteine (CYS, C). Based on this analysis, we conclude that our model provides accurate predictions for amino acid types and can be utilized for protein design.

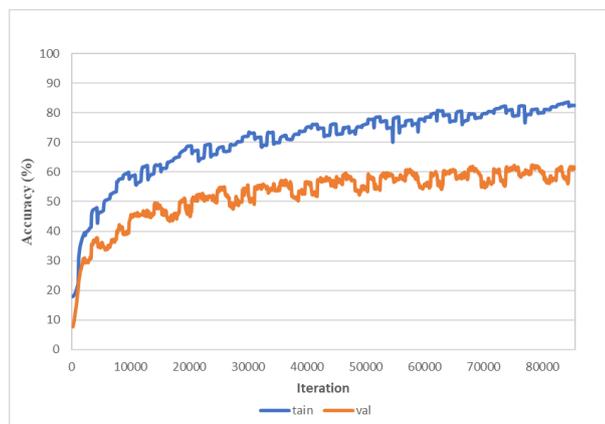

**Figure 4** shows the change in training and validation accuracy over the number of iterations.

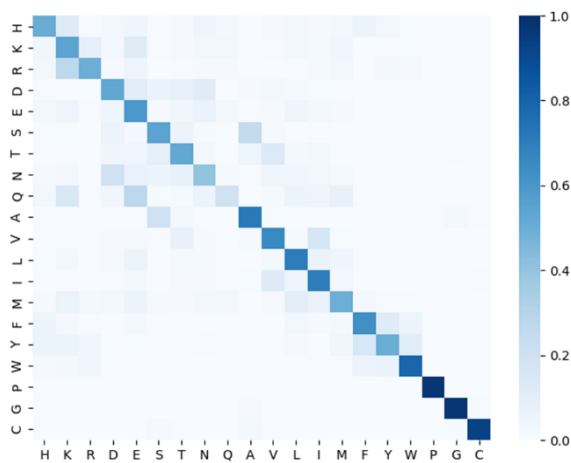

**Figure 5** presents the confusion matrix for the classification of 20 amino acids on the training set.

## 3.3 Testing Dataset Results

Table 1 presents a comparison of amino acid prediction accuracies between our method (EMOCPD) and five other methods: DenseCPD [37], ProDCoNN [36], SPROF [35], SPIN2 [34], and Wang's Model [49] on the TS500 and TS50 datasets. As shown in Table 1, our method achieves accuracies of 68.33% and 63.32% on the two datasets, respectively. Among the six methods compared, EMOCPD has the highest accuracy, exceeding DenseCPD (the second-best method) by more than 10%.

**Table 1** Comparison of accuracy between EMOCPD and other methods on the TS500 and TS50 datasets.

| Method | Test Datasets | |
| --- | --- | --- |
| | TS500 | TS50 |
| EMOCPD(ours) | **68.33%** | **62.32%** |
| DenseCPD[37] | 55.53% | 50.71% |
| ProDCoNN[36] | 42.20% | 40.69% |
| SPROF[35] | 40.25% | 39.16% |
| SPIN2 [34] | 36.60% | 33.60% |
| Wang's Model[49] | 36.14% | 33.00% |

Figure 6 presents a comparison of the TOP-K accuracy for amino acid predictions between our method and DenseCPD, ProDCoNN, and SPROF on the TS500 dataset. As shown in Figure 6, our method consistently outperforms the other methods across different K values. Specifically, our method exceeds 95% accuracy at K values of 5, whereas DenseCPD, ProDCoNN, and SPROF require K values of 9, 12, and 13, respectively, to achieve the same accuracy. According to reference [36], better TOP-K accuracy implies a smaller search space for protein design, leading to higher efficiency in designing proteins. Therefore, it can be concluded that our method has the potential for efficient protein design.

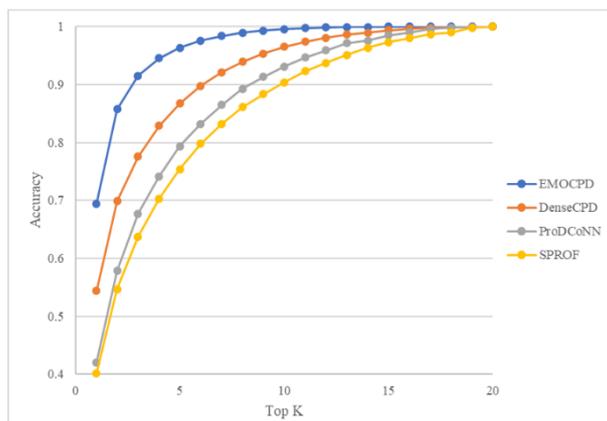

**Figure 6** Comparison of TOP-K accuracy between EMOCPD, DenseCPD, ProDCoNN, and SPROF on the TS500 dataset.

Figure 7 shows the Rec, Pre, and F1 metrics for predicting 20 amino acids using our method, DenseCPD, ProDCoNN, and SPROF on the TS500 dataset. As seen in Figure 7, our method outperforms the other three methods in both Rec and F1 metrics across all 20 amino acids, particularly excelling in the five amino acids: threonine (THR, T), arginine (ARG, R), tryptophan (TRP, W), cysteine (CYS, C), and methionine (MET, M). For the Pre metric, our method performs better than the others for 16 amino acids, while achieving comparable results for the remaining four amino acids―glutamine (GLN, Q), methionine (MET, M), asparagine (ASN, N), and glycine (GLY)―with their best-performing counterparts (GLN, MET, ASN performed best by DenseCPD; GLY performed best by SPROF). Overall, the analysis of the Rec, Pre, and F1 metrics indicates that our method demonstrates the best overall performance on the TS500 dataset.

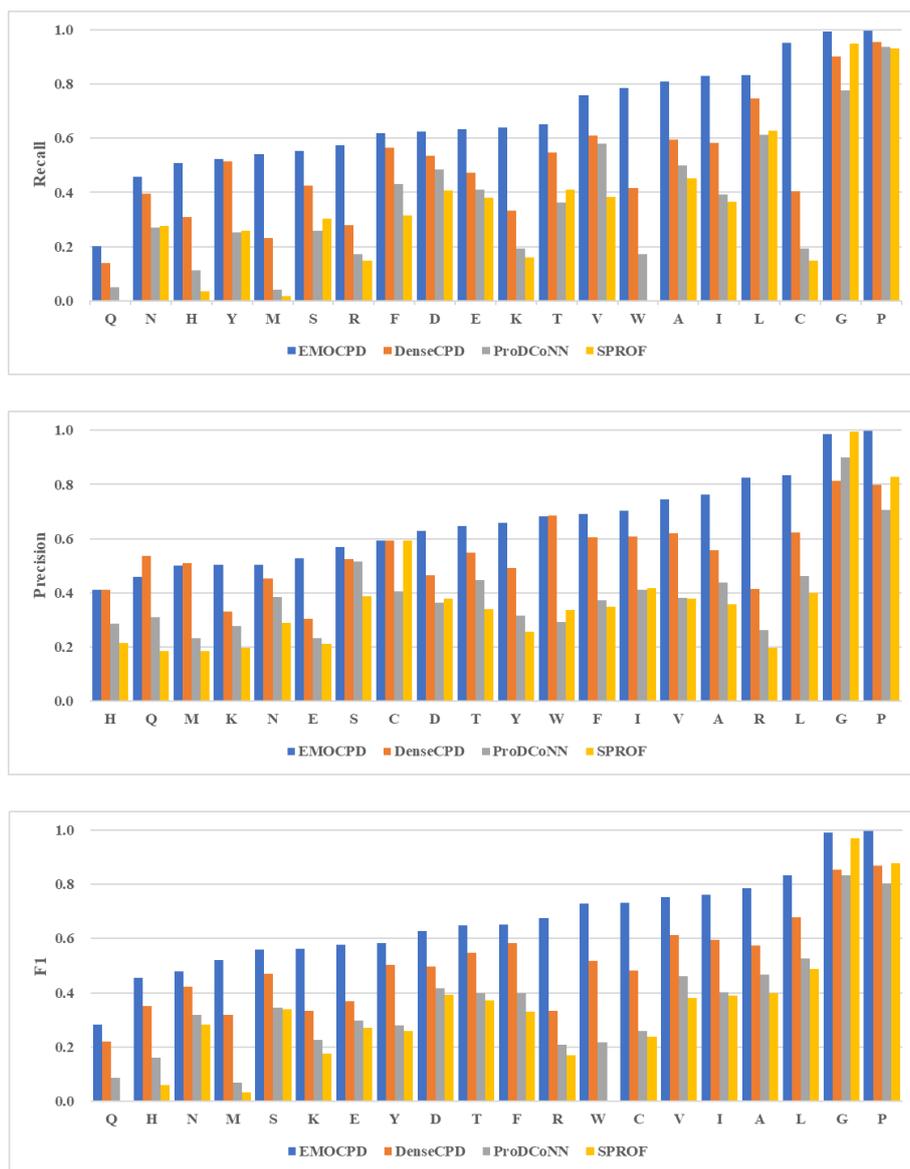

**Figure 7** Comparison of Recall, Precision, and F1 Metrics for 20 Amino Acids Using EMOCPD, DenseCPD, ProDCoNN, and SPROF on the TS500 Dataset.

### 3.4 Case Study

To clearly illustrate the advantages of our algorithm compared to others, we conducted a case study using the wild-type PETase (WTPETase, PDB ID: 5xjh). WTPETase is a single-chain protein composed of 263 amino acids, which catalyzes the hydrolysis of PET plastic. For clarity in prediction results, Figure 8 displays the amino acid sequence of WTPETase along with the predictions made by our method, EMOCPD, and DenseCPD [37]. From the figure, it can be seen that EMOCPD correctly predicted 192 amino acids with an accuracy of 73.00%, while DenseCPD correctly predicted 155 amino acids with an accuracy of 58.94%. Additionally, EMOCPD and DenseCPD correctly predicted 65 and 28 amino acids, respectively, that the other method misclassified. The number of amino acids predicted correctly by both methods is 127, while the number predicted incorrectly by both is 43. This indicates that EMOCPD provides more accurate predictions for WTPETase compared to DenseCPD.

```
EMOCPD  P A H K R G P K P T A A A L E A S A G P F A V R S F T V E N P D G F G A G T V Y Y P T S
DenseCPD G S N R L G P P P T E E S L R A P R G P F K V E S F T V P N P Q G F G A A V Y Y P T D
True    G S H M R G P N P T A A S L E A S A G P F T V R S F T V S R P S G Y G A G T V Y Y P T N

EMOCPD  A G G T V G A I S I V P G K N A R I A A I E W W G P R L A A H G F V V I T I T T E S V Q
DenseCPD P G G K V G A V A I V P G R T G T R S E V I W W G P R L A S H G F V V A A I D R K S P N
True    A G G T V G A I A I V P G Y T A R Q S S I K W W G P R L A S H G F V V I T I D T N S T L

EMOCPD  D Q P T A R A S S T M A A L R E V A S L N G N S S S P I K G K V D T S R E G V M G K S M
DenseCPD D E P E E R A E Q Q L A A L D Q V K D L N K D P S S P I Y G K I D T S R R G A M G H S L
True    D Q P S S R S S Q Q M A A L R Q V A S L N G T S S S P I Y G K V D T A R M G V M G W S M

EMOCPD  G G G G A L I S A A L N P D L M A A A P M C P W D T S T N F S S I T V P V L I L A C K D
DenseCPD G G G G S L I A A A E N P G L K A C A P M G P S H P T T D F S S I T V P T L F F A C E N
True    G G G G S L I S A A N N P S L K A A A P Q A P W D S S T N F S S V T V P T L I F A C E N

EMOCPD  D S V A P V D D T A L P I Y D S M S R V A R M F L E I N G G S H S C S T S G T S N E A L
DenseCPD D G V A P V D T H S L P M Y E S I T G N T K K Y L V I K G G H G G A E T G N D N G D L
True    D S I A P V N S S A L P I Y D S M S R N A K Q F L E I N G G S H S C A N S G N S N Q A L

EMOCPD  I G K K G V A W M K R W I E D D T R Y S T Y A C E N P D S D Q V A E Y R S S N C S L I
DenseCPD I G M K R V A W L R R H M D N D D R Y K Y F A C A D P K S D E V S E F E S K N C G L A
True    I G K K G V A W M K R F M D N D T R Y S T F A C E N P N S T R V S D F R T A N C S L E
```

**Figure 8** EMOCPD and DenseCPD Prediction Results for WTPETase. "True" represents the actual amino acid sequence; blue indicates the amino acids correctly predicted by EMOCPD; yellow indicates the amino acids correctly predicted by DenseCPD; green indicates the amino acids correctly predicted by both methods; white indicates the amino acids incorrectly predicted by both methods.

### 3.5 Biological Experimental Results

The experimental results above demonstrate the excellent theoretical performance of EMOCPD. To prove that EMOCPD can design high-quality proteins, we used it to perform site-directed mutagenesis on several key sites of the PET-degrading enzyme TfM7. The effects of these mutations were evaluated through enzyme activity assays and stability tests, as shown in Figure 9.

For example, the mutant N213T exhibited relative enzyme activities of 86.30% and 80.18% after incubation at 60℃ for 12 and 24 hours, respectively, indicating high stability. More importantly, under conditions of 70℃ for 12 hours, the relative enzyme activity of mutant N213T was 17.99%, which is 1.8 times that of the wild-type protein; after 24 hours, it showed a relative enzyme activity of 8.81%, which is 2.9 times that of the wild-type protein. This suggests that N213T has enhanced stability under high-temperature conditions. Additionally, mutants D86N and A112S also exhibited higher relative enzyme activities than the wild-type after 24 hours at both 60℃ and 70℃, indicating that the mutants predicted by our model are better suited for high-temperature, long-duration reaction conditions.

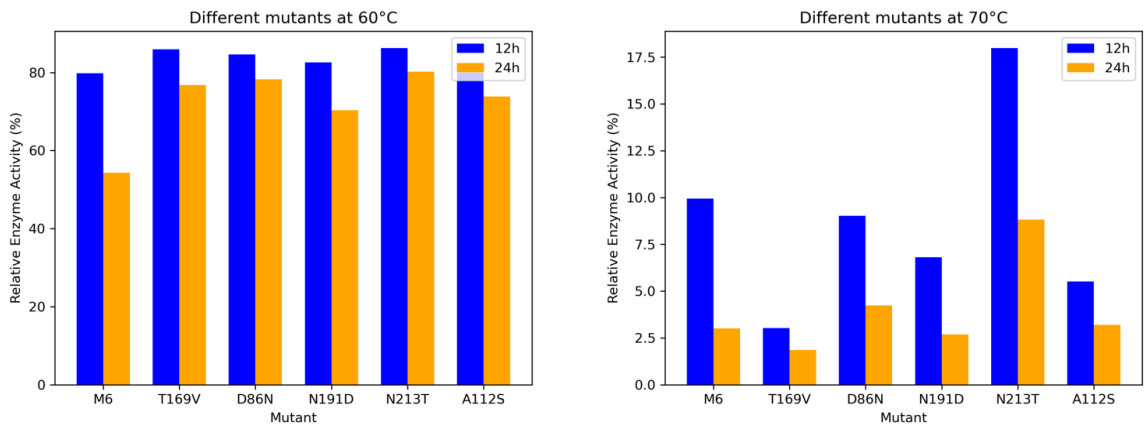

**Figure 9** Relative enzyme activities of wild-type and mutant proteins at 60℃ and 70℃.

We also assessed the protein expression levels of different mutants, and the results are shown in Figure 10. Among them, the expression level of the mutant I83M is 0.09 mg/OD · ml, which is 1.8 times that of the wild type. Additionally, the expression levels of mutants R47K and A112S also outperform the wild type. This indicates that our model can optimize the expression levels of wild-type proteins as well.

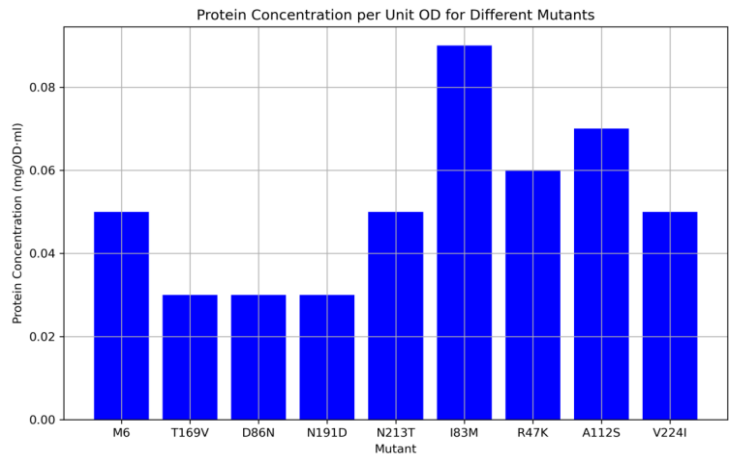

**Figure 10** Protein expression levels of different mutants.

Through the comparison of the stabilities of different mutants, we found that the EMOCPD model can identify key positions that perform exceptionally well under specific conditions. Notably, our mutants demonstrate enhanced stability and better protein expression under high-temperature conditions, which validates the effectiveness of the model's predictions.

## 4  Discussion

In previous works [36, 37], the accuracy of the proposed methods showed a strong correlation with the abundance of one or more combinations of amino acids in the predicted proteins, which may be a common characteristic of deep learning approaches. To evaluate whether our method—EMOCPD—exhibits this characteristic, this section discusses the relationship between the prediction accuracy of EMOCPD and the content of the 20 amino acids.

To investigate this relationship, we calculated the Pearson correlation coefficient and significance test probabilities between the amino acid prediction accuracy for individual protein structures and the content of the 20 amino acids. The results are shown in Table 2. From the perspective of the correlation coefficient, EMOCPD demonstrated low Pearson correlation coefficients for all amino acids, none exceeding 0.4, indicating a weak correlation between prediction accuracy and amino acid content. In terms of significance, more than half (11 amino acids) had significance probabilities less than 1%, suggesting that the prediction accuracy of EMOCPD is significantly influenced by the content of these amino acids. Among the 11 amino acids that passed the 99% significance test, 7 amino acids had negative Pearson coefficients, while 4 amino acids were positive. We categorize the former as negative amino acids and the latter as positive amino acids. The remaining 9 amino acids, which did not pass the significance test, are considered neutral amino acids. We hypothesize that a higher abundance of negative amino acids leads to lower prediction accuracy for EMOCPD, while a higher abundance of positive amino acids results in improved prediction accuracy.

**Table 2** Pearson Correlation Coefficients and Significance Probabilities Between EMOCPD Accuracy and the Content of 20 Amino Acids on the TS500 Dataset.

|   | H | K | R | D | E | S | T | N | Q | A |
|---|---|---|---|---|---|---|---|---|---|---|
| $R$ | -0.03 | -0.08 | -0.06 | 0.04 | -0.16 | -0.09 | 0.06 | -0.16 | -0.33 | 0.32 |
| $p$ | 0.34 | 0.02 | 0.08 | 0.26 | **2.5e-6** | **8e-3** | 0.07 | **3.6e-6** | **3.8e-24** | **4.8e-22** |
|   | V | L | I | M | F | Y | W | P | G | C |
| $R$ | 0.17 | -3e-3 | -0.11 | -0.20 | 0.01 | -0.07 | 0.06 | 0.20 | 0.34 | -0.20 |
| $p$ | **7.5e-7** | 0.93 | **1e-3** | **3.4e-9** | 0.76 | 0.03 | 0.08 | **3.3e-9** | **5.2e-25** | **1.6e-9** |

To verify this hypothesis, we plotted the relationship scatter plots between the accuracy of individual structures and the content of positive, negative, and neutral amino acids in the TS500 dataset, as shown in Figure 11. From the figure, it can be seen that the Pearson correlation coefficient between accuracy and the content of negative amino acids is -0.46, the correlation coefficient between accuracy and the content of positive amino acids is 0.46, and the correlation coefficient between accuracy and the content of neutral amino acids is 0.05. This result aligns with our hypothesis.

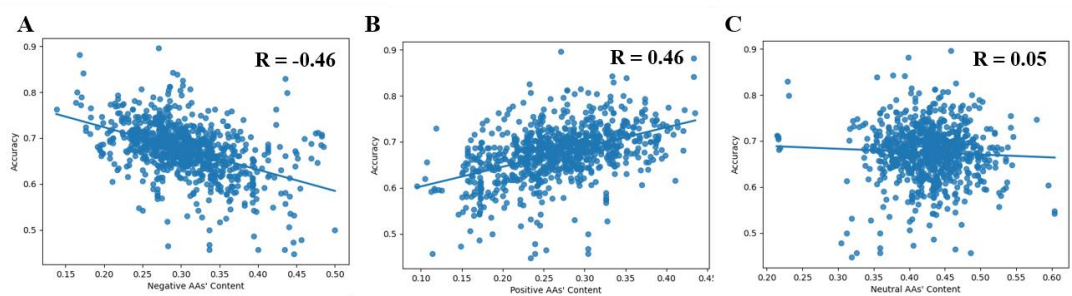

**Figure 11** Relationship between the accuracy of EMOCPD on each protein structure in TS500 and the content of negative, positive, and neutral amino acids.

Additionally, we plotted the distribution of prediction accuracy for individual structures of EMOCPD in the TS500 dataset, as shown in Figure 12. From Figure 12, it can be observed that the peak of the prediction accuracy distribution for each protein structure is around 0.68 to 0.69, which is consistent with the accuracy values reported in Table 1. The structure with the highest predicted accuracy in the TS500 dataset is 1lniA (89.58%), while the structure with the lowest predicted accuracy is 5yufA (44.68%), showing a difference of more than double. To explain this phenomenon, we provided pie charts displaying the percentages of positive, negative, and neutral amino acid content for both structures, as seen in Figure 13. From Figure 13, it is evident that the negative amino acid content in 5yufA is 45%, significantly higher than that in 1lniA (27%), which also explains the lower prediction accuracy of 5yufA, aligning with our previous hypothesis.

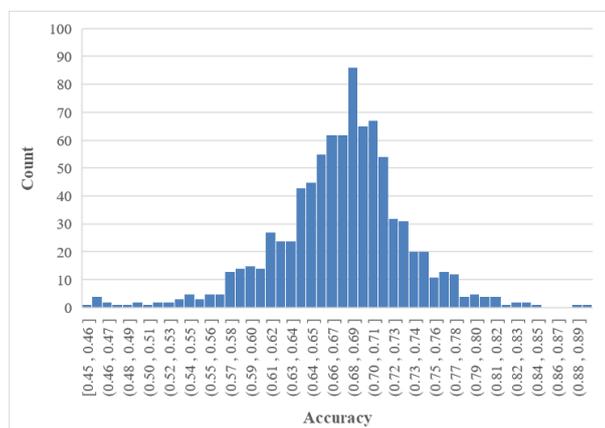

**Figure 12** Histogram of prediction accuracy distribution for each protein structure in the TS500 dataset using EMOCPD.

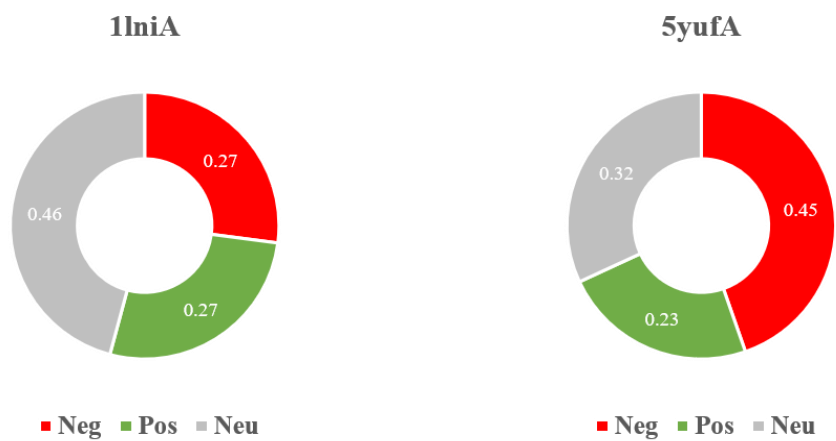

**Figure 13** Proportions of negative, positive, and neutral amino acids in proteins 1lniA and 5yufA.

# 5 Conclusion

This paper presents a novel model called EMOCPD, based on the iRMB module, designed to predict the type of each amino acid from a given protein's three-dimensional backbone, facilitating protein sequence design. We constructed a training set comprising approximately 1.6 million amino acids and a validation set with 8,000 samples to train and evaluate the performance of EMOCPD. After eight training epochs, the model achieved an accuracy of over 80% on the training samples and 62.25% on the validation set. The prediction accuracy for all amino acids, except for glutamine (GLN), exceeded 40%, demonstrating a good level of practicality for protein design.

To validate the excellent performance of EMOCPD, we compared it with five other methods on the TS50 and TS500 datasets. EMOCPD achieved accuracies of 62.32% and 68.33% on the TS50 and TS500 datasets, respectively, outperforming the second-ranked DenseCPD by more than 10%. We also compared the TOP K accuracy curves and the recall, precision, and F1 score metrics for the predictions of 20 amino acids among EMOCPD, DenseCPD, ProDCoNN, and SPROF on the TS500 dataset. EMOCPD consistently surpassed the other three methods in TOP K accuracy at various K values, indicating its potential for superior protein design. Additionally, EMOCPD outperformed the other methods in recall and F1 score metrics for all 20 amino acids, while its precision for four amino acids, including GLN, was comparable to the best-performing method.

These results validate EMOCPD's ability to predict amino acid types based on their environmental context. To further assess EMOCPD's capabilities, we predicted the PET degrading enzyme TfM7 and found through biological experiments that the model-predicted mutant N213T exhibited greater stability under extreme conditions than the wild type, while the mutant I83M showed higher protein expression than the wild type. This process confirms EMOCPD's potential for designing proteins with exceptional properties.

Additionally, we explored the impact of the amino acid composition on EMOCPD's prediction accuracy. The absolute Pearson correlation coefficients of EMOCPD's prediction accuracy with the contents of the 20 amino acids did not exceed 0.4, indicating a weak correlation with the composition of different amino acids. However, among the 20 amino acids, 11 amino acids passed the 99% significance test, with seven having negative Pearson coefficients and four positive. We categorized the amino acids that did not pass the significance test as neutral, those with negative coefficients as negative amino acids, and those with positive coefficients as positive amino acids. The calculated correlation coefficients for the proportions of positive, negative, and neutral amino acids relative to prediction accuracy were -0.46, 0.46, and 0.05, respectively.

This suggests that a higher content of positive amino acids correlates with increased prediction accuracy, while a higher content of negative amino acids correlates with decreased accuracy. From a protein design perspective, EMOCPD suggests that positive amino acids are less prone to mutation, while negative amino acids are more susceptible to mutation.

It is important to note that when using EMOCPD for protein design, multiple sites often need to be mutated in combination to achieve better-performing mutant proteins. Therefore, despite EMOCPD's demonstrated predictive capabilities and potential for designing excellent proteins, the strategy for using EMOCPD in protein design requires further investigation.

## References

[1] Jiang L, Althoff E A, Clemente F R, et al. De novo computational design of retro-aldol enzymes. *Science* **2008**;319:1387-1391.

[2] Röthlisberger D, Khersonsky O, Wollacott A M, et al. Kemp elimination catalysts by computational enzyme design. *Nature* **2008**;453:190-195.

[3] Siegel J B, Zanghellini A, Lovick H M, et al. Computational design of an enzyme catalyst for a stereoselective bimolecular Diels-Alder reaction. *Science* **2010**;329:309-313.

[4] Correia B E, Bates J T, Loomis R J, et al. Proof of principle for epitope-focused vaccine design. *Nature* **2014**;507:201-206.

[5] Marcandalli J, Fiala B, Ols S, et al. Induction of potent neutralizing antibody responses by a designed protein nanoparticle vaccine for respiratory syncytial virus. *Cell* **2019**;176:1420-1431.

[6] Schulte D, Peng W, Snijder J. Template-based assembly of proteomic short reads for de novo antibody sequencing and repertoire profiling. *Analytical Chemistry* **2022**;94:10391-10399.

[7] Baran D, Pszolla M G, Lapidoth G D, et al. Principles for computational design of binding antibodies. *Proceedings of the National Academy of Sciences* **2017**;114:10900-10905.

[8] Lu P, Min D, DiMaio F, et al. Accurate computational design of multipass transmembrane proteins. *Science* **2018**;359:1042-1046.

[9] Lalaurie C J, Dufour V, Meletiou A, et al. The de novo design of a biocompatible and functional integral membrane protein using minimal sequence complexity. *Scientific Reports* **2018**;8:14564.

[10] Chevalier A, Silva D A, Rocklin G J, et al. Massively parallel de novo protein design for targeted therapeutics. *Nature* **2017**;550:74-79.

[11] Silva D A, Yu S, Ulge U Y, et al. De novo design of potent and selective mimics of IL-2 and IL-15. *Nature* **2019**;565:186-191.

[12] Glasgow A A, Huang Y M, Mandell D J, et al. Computational design of a modular protein sense-response system. *Science* **2019**;366:1024-1028.

[13] Lu H, Diaz D J, Czarnecki N J, et al. Machine learning-aided engineering of hydrolases for PET depolymerization. *Nature* **2022**;604:662-667.

[14] Shroff R, Cole A W, Diaz D J, et al. Discovery of novel gain-of-function mutations guided by structure-based deep learning. *ACS synthetic biology* **2020**;9:2927-2935.

[15] Anfinsen C B. Principles that govern the folding of protein chains. *Science* **1973**;181: 223-230.

[16] Baek M, DiMaio F, Anishchenko I, et al. Accurate prediction of protein structures and interactions using a three-track neural network. *Science* **2021**;373:871-876.

[17] Jumper J, Evans R, Pritzel A, et al. Highly accurate protein structure prediction with AlphaFold. *Nature* **2021**;596:583-589.

[18] Leman J K, Weitzner B D, Lewis S M, et al. Macromolecular modeling and design in Rosetta: recent methods and frameworks. *Nature methods* 2020;17:665-680.

[19] Huang X, Pearce R, Zhang Y. EvoEF2: accurate and fast energy function for computational protein design. *Bioinformatics* 2020;36:1135-1142.

[20] Huang B, Xu Y, Hu X, et al. A backbone-centred energy function of neural networks for protein design. *Nature* **2022**;602:523-528.

[21] Rocklin G J, Chidyausiku T M, Goreshnik I, et al. Global analysis of protein folding using massively parallel design, synthesis, and testing. *Science* **2017**;357:168-175.

[22] Dosovitskiy A, Beyer L, Kolesnikov A, et al. An image is worth 16x16 words: Transformers for image recognition at scale. *arXiv preprint* **2020**;arXiv:2010.11929.

[23] Liu Z, Lin Y, Cao Y, et al. Swin transformer: Hierarchical vision transformer using shifted windows. *Proceedings of the IEEE/CVF international conference on computer vision* **2021**:10012-10022.

[24] Liu Z, Mao H, Wu C Y, et al. A convnet for the 2020s. *Proceedings of the IEEE/CVF conference on computer vision and pattern recognition* **2022**:11976-11986.

[25] Vaswani A, Shazeer N, Parmar N, et al. Attention is all you need. *Advances in neural information processing systems* **2017**;30.

[26] Radford A, Narasimhan K, Salimans T, et al. Improving language understanding by generative pre-training. **2018**.

[27] Hessel J, Marasović A, Hwang J D, et al. Do androids laugh at electric sheep? humor" understanding" benchmarks from the new yorker caption contest. *arXiv preprint* **2022**;arXiv:2209.06293.

[28] Mater A C, Coote M L. Deep learning in chemistry. *Journal of chemical information and modeling* **2019**;59:2545-2559.

[29] Goh G B, Hodas N O, Vishnu A. Deep learning for computational chemistry. *Journal of computational chemistry* **2017**;38:1291-1307.

[30] Webb S. Deep learning for biology. *Nature* **2018**;554:555-557.

[31] Wu Q, Deng Z, Pan X, et al. MDGF-MCEC: a multi-view dual attention embedding model with cooperative ensemble learning for CircRNA-disease association prediction. *Briefings in Bioinformatics* **2022**;23:bbac289.

[32] Wang Z, Deng Z, Zhang W, et al. MMSMAPlus: a multi-view multi-scale multi-attention embedding model for protein function prediction. *Briefings in Bioinformatics* **2023**:bbad201.

[33] Li Z, Yang Y, Faraggi E, et al. Direct prediction of profiles of sequences compatible with a protein structure by neural networks with fragment‐based local and energy‐based nonlocal profiles. *Proteins: Structure, Function, and Bioinformatics* **2014**;82:2565-2573.

[34] O'Connell J, Li Z, Hanson J, et al. SPIN2: Predicting sequence profiles from protein structures using deep neural networks. *Proteins: Structure, Function, and Bioinformatics* **2018**;86: 629-633.

[35] Chen S, Sun Z, Lin L, et al. To improve protein sequence profile prediction through image captioning on pairwise residue distance map. *Journal of chemical information and modeling* **2019**;60:391-399.

[36] Zhang Y, Chen Y, Wang C, et al. Prodconn-protein design using a convolutional neural network. *Biophysical Journal* **2020**;118:43a-44a.

[37] Qi Y, Zhang J Z H. DenseCPD: improving the accuracy of neural-network-based computational protein sequence design with DenseNet. *Journal of chemical information and modeling* **2020**;60:1245-1252.

[38] Huang G, Liu Z, Van Der Maaten L, et al. Densely connected convolutional networks. *Proceedings of the IEEE conference on computer vision and pattern recognition* **2017**: 4700-4708.

[39] Ingraham J, Garg V, Barzilay R, et al. Generative models for graph-based protein design. *Advances in neural information processing systems* **2019**;32.

[40] Dauparas J, Anishchenko I, Bennett N, et al. Robust deep learning–based protein sequence design using ProteinMPNN. *Science* **2022**;378:49-56.

[41] Zhang J, Li X, Li J, et al. Rethinking mobile block for efficient attention-based models. *Proceedings of*

*the IEEE/CVF International Conference on Computer Vision* **2023**:1389-1400.

[42] Burley S K, Berman H M, Kleywegt G J, et al. Protein Data Bank (PDB): the single global macromolecular structure archive. *Protein crystallography: methods and protocols* **2017**: 627-641.

[43] Dolinsky T J, Czodrowski P, Li H, et al. PDB2PQR: expanding and upgrading automated preparation of biomolecular structures for molecular simulations. *Nucleic acids research* **2007**;35:W522-W525.

[44] Mitternacht S. FreeSASA: An open source C library for solvent accessible surface area calculations. *F1000Research* **2016**;5.

[45] He K, Zhang X, Ren S, et al. Deep residual learning for image recognition. *Proceedings of the IEEE conference on computer vision and pattern recognition* **2016**:770-778.

[46] Hu J, Shen L, Sun G. Squeeze-and-excitation networks. *Proceedings of the IEEE conference on computer vision and pattern recognition* **2018**:7132-7141.

[47] Paszke A, Gross S, Massa F, et al. Pytorch: An imperative style, high-performance deep learning library. *Advances in neural information processing systems* **2019**;32.

[48] Kingma D P, Ba J. Adam: A method for stochastic optimization. *arXiv preprint* **2014**;arXiv:1412.6980.

[49] Wang J, Cao H, Zhang J Z H, et al. Computational protein design with deep learning neural networks. *Scientific reports* **2018**;8:1-9.